\documentclass[10pt,twocolumn,letterpaper]{article}

\usepackage{cvpr}
\usepackage{times}
\usepackage{epsfig}
\usepackage{graphicx}
\usepackage{amsmath}
\usepackage{ifthen} 
\usepackage{amssymb}

\usepackage[pagebackref=true,breaklinks=true,letterpaper=true,colorlinks,bookmarks=false]{hyperref}

\def\version{final} 

\ifthenelse{\equal{\version}{draft}} {
	\def\ienote#1{{\color{red}[Irfan:#1]}} 
	\def\dcnote#1{{\color{blue}[Daniel:#1]}}
	\def\jhnote#1{{\color{green}[James:#1]}}
	\def\shnote#1{{\color{magenta}[Steven:#1]}}
	\def\psnote#1{{\color{cyan}[Patsorn:#1]}}
	\def\TODOnote#1{{\color{red}[TODO:#1]}}
}
{
 	\def\ienote#1{} 
 	\def\dcnote#1{}
 	\def\jhnote#1{}
 	\def\shnote#1{}
 	\def\psnote#1{}
 	\def\TODOnote#1{}
}

\cvprfinalcopy 

\def\cvprPaperID{3034} 

\begin{document}
\title{Let's Dance: Learning From Online Dance Videos}
\author{Daniel Castro\\
Georgia Institute of Technology\\
{\tt\small dcastro9@gatech.edu}
\and
Steven Hickson\\
{\tt\small shickson@gatech.edu}
\and
Patsorn Sangkloy\\
{\tt\small patsorn\_sangkloy@gatech.edu}
\and
Bhavishya Mittal\\
{\tt\small bmittal6@gatech.edu}
\and
Sean Dai\\
{\tt\small sdai@gatech.edu}
\and
James Hays\\
{\tt\small hays@gatech.edu}
\and
Irfan Essa\\
{\tt\small irfan@gatech.edu}
}

\maketitle

\begin{abstract}
In recent years, deep neural network approaches have naturally extended to the video domain, in their simplest case by aggregating per-frame classifications as a baseline for action recognition. A majority of the work in this area extends from the imaging domain, leading to visual-feature heavy approaches on temporal data. To address this issue we introduce ``Let's Dance'', a 1000 video dataset (and growing) comprised of 10 visually overlapping dance categories that require motion for their classification. We stress the important of human motion as a key distinguisher in our work given that, as we show in this work, visual information is not sufficient to classify motion-heavy categories. We compare our datasets' performance using imaging techniques with UCF-101 and demonstrate this inherent difficulty.  We present a comparison of numerous state-of-the-art techniques on our dataset using three different representations (video, optical flow and multi-person pose data) in order to analyze these approaches. We discuss the motion parameterization of each of them and their value in learning to categorize online dance videos. Lastly, we release this dataset (and its three representations) for the research community to use.
\end{abstract}

\section{\label{sec:intro}Introduction}

Video is a rich medium with dynamic information that can be used to determine, what is happening in a scene. In this work, we consider \textsl{highly dynamic video}, video that requires the parametrization of motion over extended sequences to identify the activity being performed. The main challenge with highly dynamic video is that a single frame cannot provide sufficient information to understand the action being performed. Therefore, multiple frames, leading to an extended sequence of frames, need to be analyzed for scene classification. One of the drawbacks of current action classification research is both a lack of approaches that can be applied to extended/long sequences and datasets lacking in such highly dynamic videos. Our goal is to determine which methods best represent motion as opposed to methods that use a single (properly picked) frame ~\cite{feichtenhofer2016convolutional} to identify the activity, as we feel such approaches devalue the necessity for video data. In this work we introduce a 1,000 video dataset and evaluate methods that focuses on highly dynamic videos requiring motion analysis for classification.

We choose the domain of dance videos as (a) there is large amount of dance videos available online and (b) the diversity of dynamics in these videos provides us with a challenging space of problems for highly dynamic video analysis. This enables us to conduct a focused study on the relevance of motion in dancing classification and the broader value of motion in improving video classification.

The core challenge of this task is attaining an adequate representation of human motion across a 10-second clip. In order to highlight the trajectory of this work, we will evaluate the current approaches and demonstrate the value of isolating motion for properly evaluating these approaches and this dataset.

\begin{figure}[t]
	\includegraphics[width=0.5\textwidth]{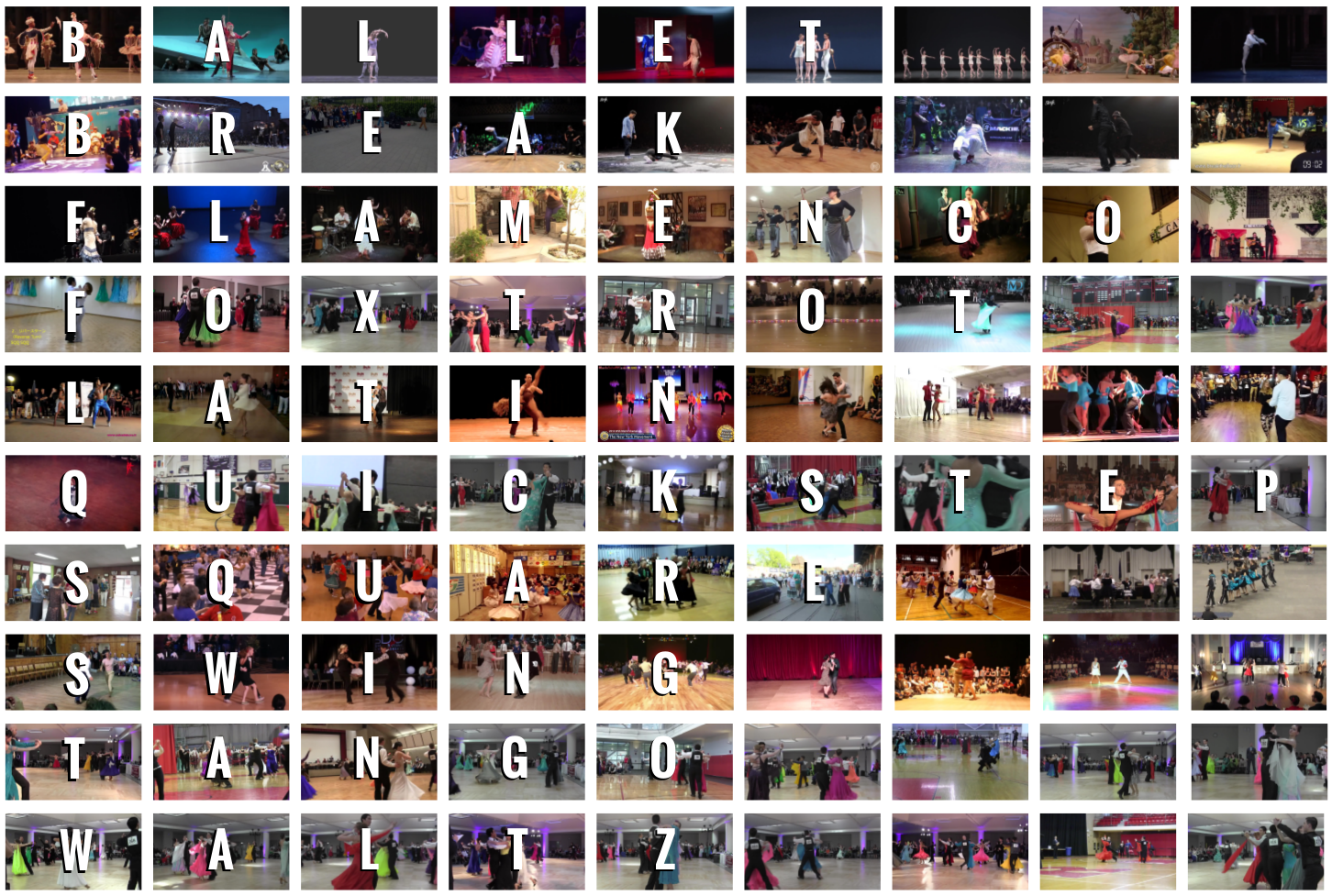}
	\caption{\label{fig:videos}Each row contains frames from the class it represents. This figure is best viewed digitally.}
\end{figure}
Many video classification techniques exist, either utilizing single frames, late fusion architectures, temporal (3D) convolutional networks, or recurrent networks with long short-term memory (LSTM). Current classification problems can often be identified by a single frame. We present a more challenging problem wherein each class requires the use of multiple frames to adequately classify each category.

Specifically, we propose the use of optical flow and multi-agent pose estimation as motion representations which augment traditional video classification approaches. Comparing these approaches enables us to gain insights into the inherent encoding of motion in neural networks that is difficult to understand.

Our main contributions are: \textbf{(1)} An analysis of baseline and state-of-the-art approaches in video classification, \textbf{(2)} a general method for concurrently learning from multiple motion parameterizations in video, and \textbf{(3)} A 1000 video dataset of highly dynamic dance videos, contrasted with existing video datasets, to motivate further investigation and understanding of motion parameterization in video classification.

\section{\label{sec:related}Related Work}

In order to determine which competing state-of-the-art approaches to examine, we first present a literature review on video classification. While deep networks have been shown to be very effective at classifying, localizing, and segmenting images, it is still unclear how to properly extend these methods to the video domain. There are a few common approaches, some of which are: (1) Applying proven image classification deep network architectures to individual frames of a video; (2) Extending 2D convolutional operators to 3D convolutions acting on the time domain;  and (3) Preprocessing the video into images that encode motion, such as optical flow, and running current image architectures on the processed frames.

A simple way to extend image-based neural networks to video classification is to extract features from each individual frame of a video~\cite{NIPS2012_4730}. While this technique does lead to some success if the network learns temporally-invariant features, it is commonly only used as a baseline approach to compare against networks that incorporate temporal data~\cite{6165309,Karpathy_2014_CVPR}. One common variant is a two-stream late fusion architecture with a still frame-based ``spatial'' network stream running in parallel alongside a ``temporal'' network performing classifications based on optical flow calculations~\cite{cheron2015p,feichtenhofer2016convolutional,gkioxari2015finding,NIPS2014_5353}. This network architecture significantly outperforms approaches based solely on individual frame classification, suggesting that incorporating a temporal component is necessary. In our work we leverage the benefit of a temporal network by incorporating it into the design of our network architecture.

Karpathy et al.~explore more direct methods of incorporating temporal data with each video frame by extending the convolution kernels from size $m$ x $m$ x $3$ to $m$ x $m$ x $3$ x $T$, where $T$ represents a temporal extent\cite{Karpathy_2014_CVPR}. They also point out one of the major challenges of using deep learning for video classification -- there are no large-scale video datasets comparable to the quality and size of image recognition datasets. Similarly, 3D convolutional kernels that incorporate the spatial domain have been shown to be successful for action classification in both security camera and depth data recordings \cite{6165309,Wang:2014:HAR:2647868.2654912}. Wang et al.~use a similar two-stream late fusion approach \cite{Wang_2015_CVPR}, but they note that without incorporating the learned features into an ensemble method with handcrafted features, these deep-learned approaches still fail to outperform handcrafted approaches. We combine these methods in our work by incorporating preprocessed features (optical flow and multi-agent pose detection) with 3D convolutional kernels in order to integrate the representation of motion into the network architecture.

Another common approach is to leverage the sequential nature of a Long Short-Term Memory (LSTM) network--a specific type of recurrent neural network with additional gates to control the flow of information. LSTMs can process information over long term temporal sequences and have been applied in video for various tasks such as caption generation \cite{venugopalan2015sequence} and learning video representations \cite{DBLP:journals/corr/SrivastavaMS15}. Similarly, the long-term recurrent convolutional networks (LRCNs) proposed by Donahue et. al. introduce another variation of an LSTM for this task. Despite their temporal nature, these approaches have been less successful at encoding motion in comparison to two-stream networks \cite{donahue2015long} which encode the spatial and temporal domain in concurrent architectures.

The most effective method for classifying motion in video is still unclear. In the context of action recognition, many of these approaches are learning features based on the image's context and not the inherent action. This is in part because commonly used video datasets such as UCF-Sports and more traditionally UCF-101 can generally be identified to moderately decent accuracy using single-frame approaches which do not encode motion parameters\cite{Karpathy_2014_CVPR}.

A specific method for encoding motion that has recently gained traction in action recognition is the use of pose detection over the temporal domain with neural networks \cite{cheron2015p}\cite{liu2016spatio}\cite{toshev2014deeppose}\cite{wei2016convolutional}. Detecting pose over this domain provides us with the intrinsic motion of the subjects in the scene. As highlighted earlier, an initial breakthrough was achieved by Toshev et. al.~\cite{toshev2014deeppose} with state-of-the-art results in estimating the pose of a single individual from an image. The importance of pose was further demonstrated in \cite{cheron2015p}, incorporating pose features from a CNN into action recognition. This work was extended over the next two years to attain joint-specific networks that work well with partial and occluded poses \cite{wei2016convolutional}. It was then most recently implemented to detect multiple people within a single frame \cite{cao2016realtime}. In our work we will leverage our own implementation of multi-agent pose detection to demonstrate the need for motion parameterization when classifying highly dynamic video.
\begin{figure}[t]
	\includegraphics[width=0.5\textwidth]{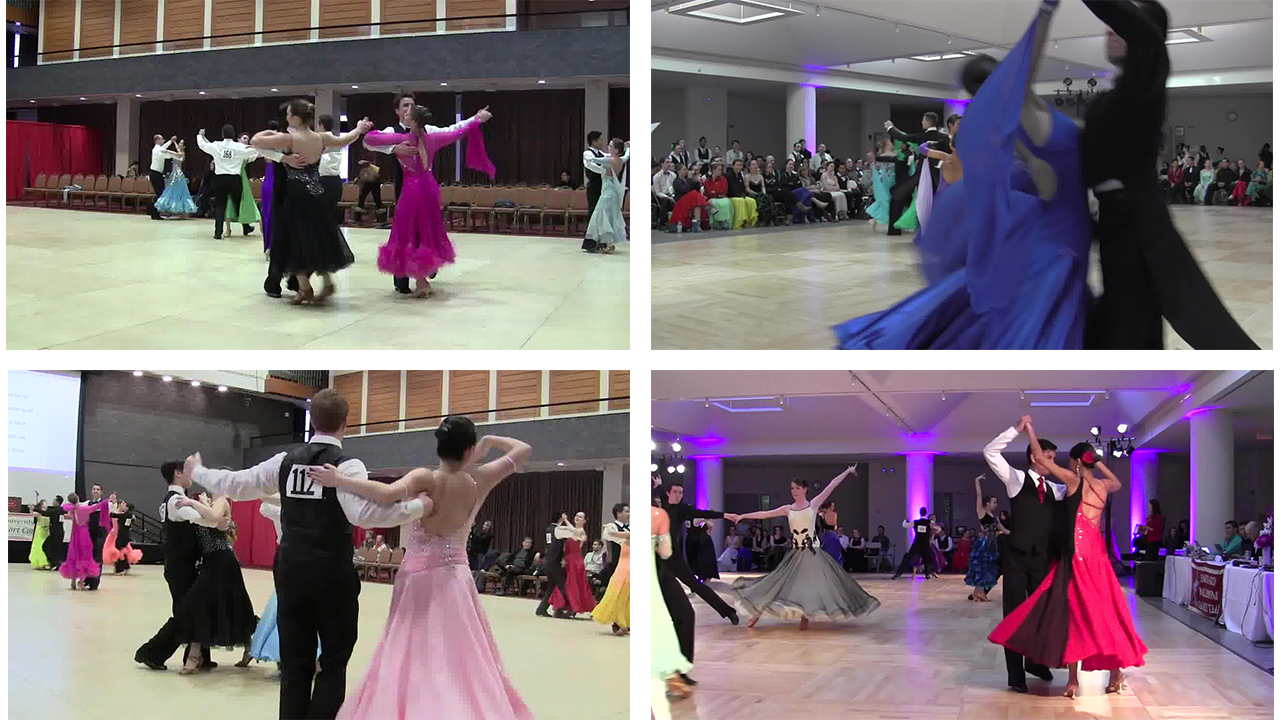}
	\caption{\label{fig:ballroom}Each of these examples represents a different class in our dataset (they are types of ballroom dancing). Top Left: Waltz, Top Right: Quickstep, Bottom Left: Foxtrot, Bottom Right: Tango.}
\end{figure}

\subsection{Existing Datasets}

There are a handful of relevant datasets that exist in the research domain. We highlight some of the more relevant video datasets that are appropriate to our work. All of these datasets demonstrate the growing need for understanding what type of motion features are relevant in classifying highly dynamic actions, which we explore in this work.

\subsubsection{UCF-101}

The UCF-101 dataset \cite{soomro2012ucf101} contains approximately 13,000 clips and 101 action classes, totaling 27 hours of data. The clip length varies largely from ~1 second to 71 seconds depending on the activity at a resolution of 320x240. This was one of the first datasets to tackle human actions in video. However, as we will demonstrate in this work, most per-frame (image-based) approaches still perform moderately well on the dataset, illustrating the main question which we seek to answer in this work -- that being the representation of motion as a classification feature.

\subsubsection{Kinetics}

The Kinetics dataset \cite{kinetics} contains 300,000 clips and 400 action classes, with a minimum of 400 videos per class. The action classes are also loosely grouped in 32 parent classes which further break down the dataset. This dataset was collected semi-automatically with curation through image classifiers and use of Amazon Mechanical Turk to determine the action classes and if the video snippet was appropriate to that class.

\subsubsection{Atomic Visual Actions (AVA)}

The AVA dataset \cite{ava} contains 80 atomic visual actions in 57,400 movie clips which are localized within the frame. This work goes beyond simply understanding a simple action in a video clip to understanding the interaction, both between humans and with humans and objects. Although this is somewhat less relevant to our work, it demonstrates the need for understanding motion features in human interaction -- specifically by localizing the action and its relevance in a scene that may contain multiple subjects / objects.

\section{Let's Dance Dataset}

Our main challenge in this work was determining a reliable way of testing how well a specific method can parameterize motion. We realized that available video datasets such as UCF-101 \cite{soomro2012ucf101} and UCF-Sports \cite{rodriguez2008action} \ienote{Citations?} tackled a known classification problem, one that could be evaluated using extensions of available image classification architectures.

With that in mind, we developed a new dataset that prioritizes motion as the key characteristic of the classification.  We assembled a 1,000 video dataset containing 10 dynamic and visually overlapping dances. We chose the parent category of dancing because it has a variety of measurable features (\ie\ rhythm, limb movement), and it is not represented in the Sports-1M and UCF-101 datasets~\cite{Karpathy_2014_CVPR, soomro2012ucf101}.
\
The categories included for this dataset are:
\begin{center}
\begin{tabular}{ll}
\textbullet~Ballet & \textbullet~Break Dancing \\
\textbullet~Flamenco & \textbullet~Foxtrot \\
\textbullet~Latin & \textbullet~Quickstep \\
\textbullet~Square & \textbullet~Swing \\
\textbullet~Tango & \textbullet~Waltz \\
\end{tabular}
\end{center}

The dataset contains 100 videos for each class. Each video is 10-seconds long at 30 frames per second. The videos themselves were taken from YouTube at a quality of 720p, and includes both dancing performances and plain-clothes practicing.  Examples of each class can be seen in Figure~\ref{fig:videos}.

We highlight that the dataset contains four different types of ballroom dancing (quickstep, foxtrot, waltz, and tango) as seen in Figure \ref{fig:ballroom}. The motivation behind picking these dances is that their parent category is specifically the setting in which the dance occurs (a ballroom). This satisfies our main challenge of selecting classes that exemplify highly dynamic video. On this note, we extract two different motion representations from our input data for use by the community; optical flow\cite{farneback2003two} and multi-person pose detection.

\begin{figure}[t]
\centering
\includegraphics[width=0.5\textwidth]{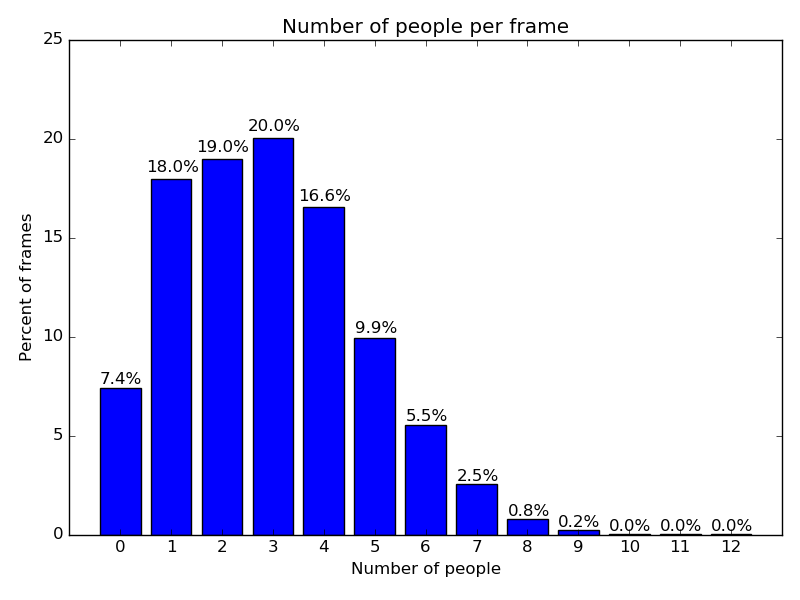}
\caption{Distribution of number of people per frame using \cite{redmon2015you}. 75\% of frames had at least two people detected in the dataset. 56\% of the dataset has more than two people in the shot, which further illustrates the added complexity of this task.}
\label{fig:skeldist}
\end{figure}

When attempting to detect pose, we found numerous methods that focused on single-person pose detection. We adapted these methods to multiple individuals (given that dancing is generally a group activity, see Figure \ref{fig:skeldist}) through the use of a recent real-time person detector\cite{redmon2015you}. Similar approaches can be seen in  \cite{gkioxari2014using}\cite{pishchulin2012articulated}\cite{cao2016realtime}.

After detecting the bounding boxes for each person in the scene we computed the pose for each individual using \cite{wei2016convolutional}. Positive and negative examples of this methodology can be seen in Figure \ref{fig:skeletons}.


\section{\label{sec:baselinemethods}Baseline Methods}
In order to better understand the need for motion parametrization in video, we have identified two commonly-used architectures to establish as our baseline. These are architectures which are commonly applied to video architectures but only take a single-frame as input (per architecture). 

These approaches are extensions of very successful image classification techniques. 

\begin{figure}[b]
\centering
\includegraphics[width=0.5\textwidth]{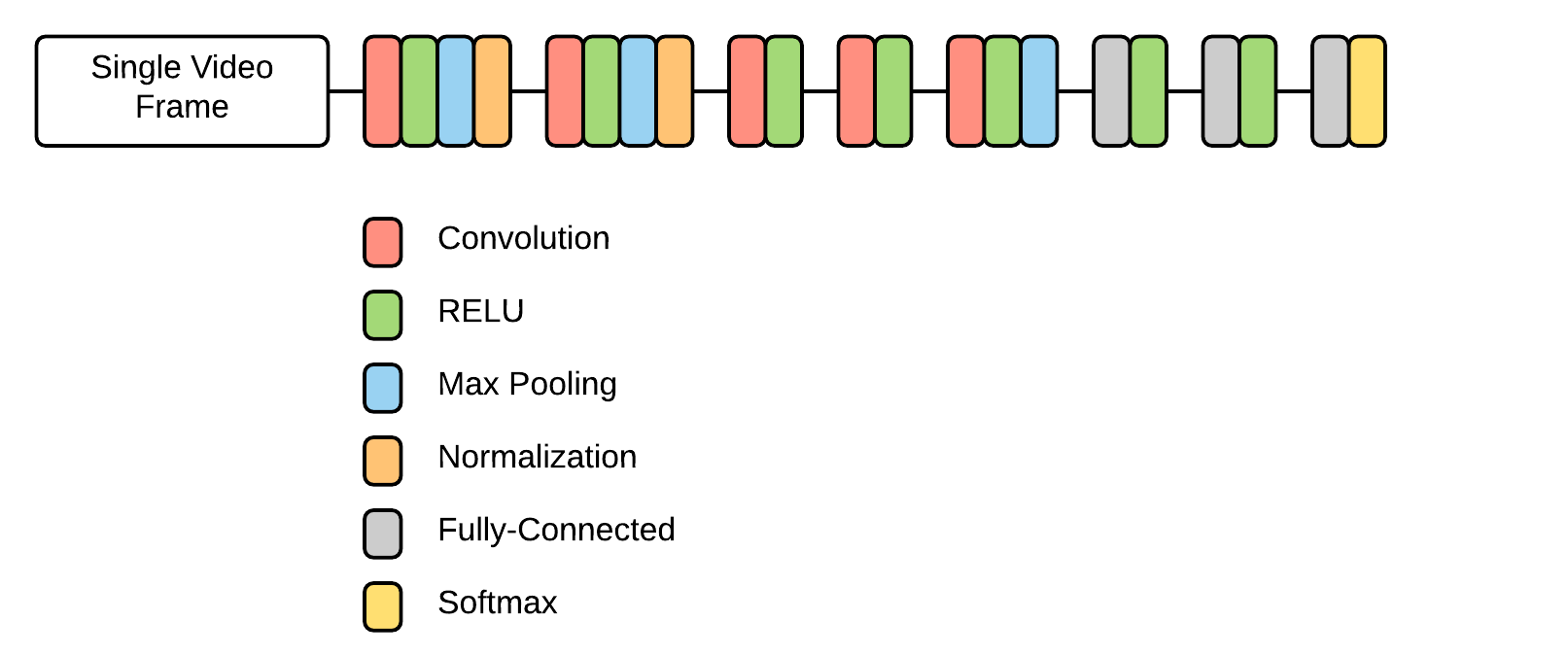}
\caption{Frame-by-Frame Architecture: This is a traditional CNN, commonly used in image recognition.}
\label{fig:arch1}
\end{figure}

\subsection{Frame-by-Frame} Using the architecture of a state-of-the-art convolutional neural network for image classification, such as VGG \cite{simonyan2014very}, a classification for the video can be achieved based on key image frames from a video.  A sample architecture based on CaffeNet, a variation of AlexNet \cite{NIPS2012_4824}, is shown in Figure \ref{fig:arch1}. This approach does not explicitly encode motion in determining the video's classification but rather categorizes each frame and naively selects the majority label.

We do note that although there are numerous approaches for aggregating a single class from multiple per-frame classifications, the network itself does not encode the temporal domain.

\subsection{Two-Stream Late Fusion} 
\begin{figure}[htp!]
\centering
\includegraphics[width=0.47\textwidth]{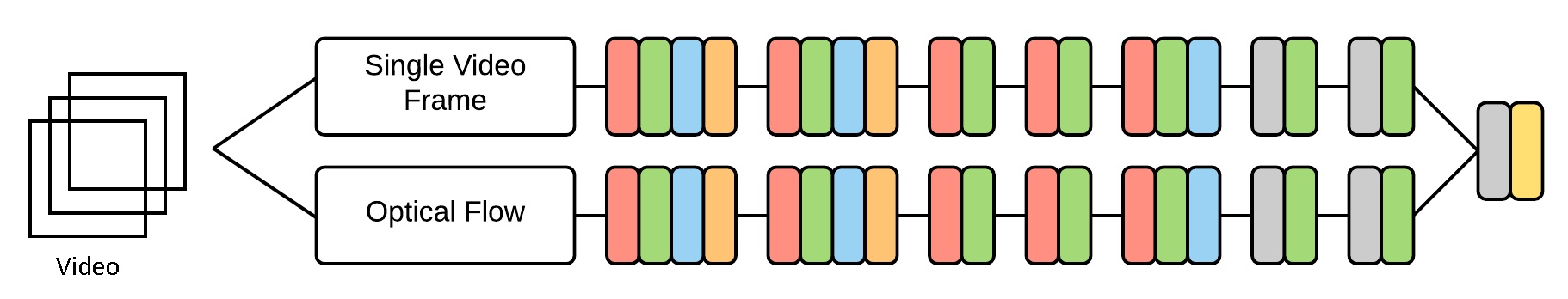}
\caption{Two-Stream Late Fusion Architecture (color key in \ref{fig:arch1}). This method incorporates motion (optical flow) into a traditional CNN pipeline.}
\label{fig:arch2}
\end{figure}

A common way of adding a temporal component to deep networks is by separately performing a classification based on spatial data (a single frame) and temporal data (i.e. optical flow). Merging these results produces an overall classification for the video, as shown in Figure \ref{fig:arch2}.

This approach computes optical flow from two frames (at time $n$ and $n-k$ where $k$ is not necessarily 1) over the period of the entire video. Each frame in this case can be considered a single instance of motion that occurred in the video. For dancing we envision this as a specific move in a dance.

\section{Proposed Approaches}

In order to address the challenge of categorizing highly dynamic videos we implement a number of methods which explicitly encode motion. At the core of these approaches is the notion of 3-dimensional kernels which process a series of video frames for classification. This enables us to pass in very short video clips (16 frames or approx. 1/2 second) for the network to learn. The overall objective was to incorporate motion in the learning pipeline of standard approaches and assess their performance. After testing these approaches it was evident that combining numerous motion parameterizations in a concurrent deep network architecture would best represent the input video. 

\subsection{Temporal 3D CNN (RGB)} As stated, traditional convolutional neural networks can be extended to video by using 3-dimensional kernels that convolve in the temporal domain. We focus on testing this slow-fusion approach discussed in \cite{6165309}, which embeds the high-level spatial and temporal information at the initial convolutional layers by propagating the information through the network. One of the main setbacks of this proposed approach is the computational time it currently takes to compute these methodologies. We discuss this further in Section \ref{results}.

\begin{figure}[t]
\centering
\includegraphics[width=0.48\textwidth]{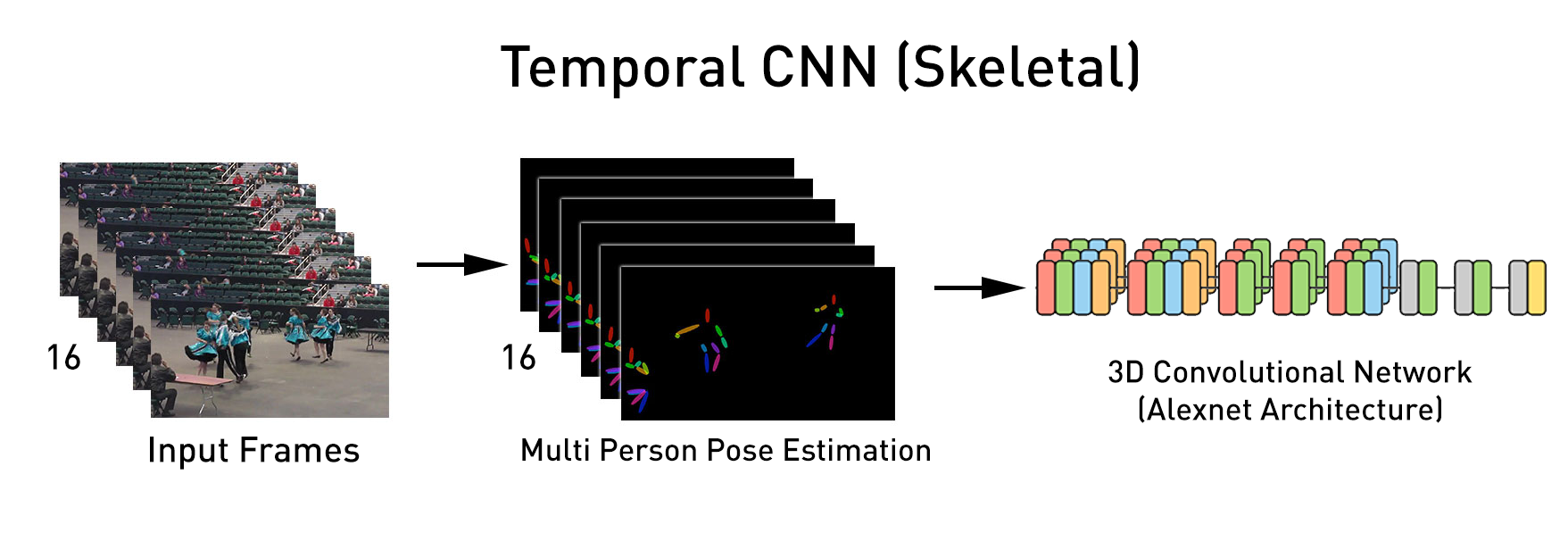}
\caption{This pipeline displays a skeletal temporal CNN (3D Convolution) which processes the initial frames to obtain a multi person pose estimation from the input frames obtained by performing a bounding box person detection from \cite{redmon2015you} which is then processed by \cite{wei2016convolutional} for detecting the dancers' pose.}
\label{fig:skeletal}
\end{figure}

\begin{figure}[t]
\centering
\includegraphics[width=0.5\textwidth]{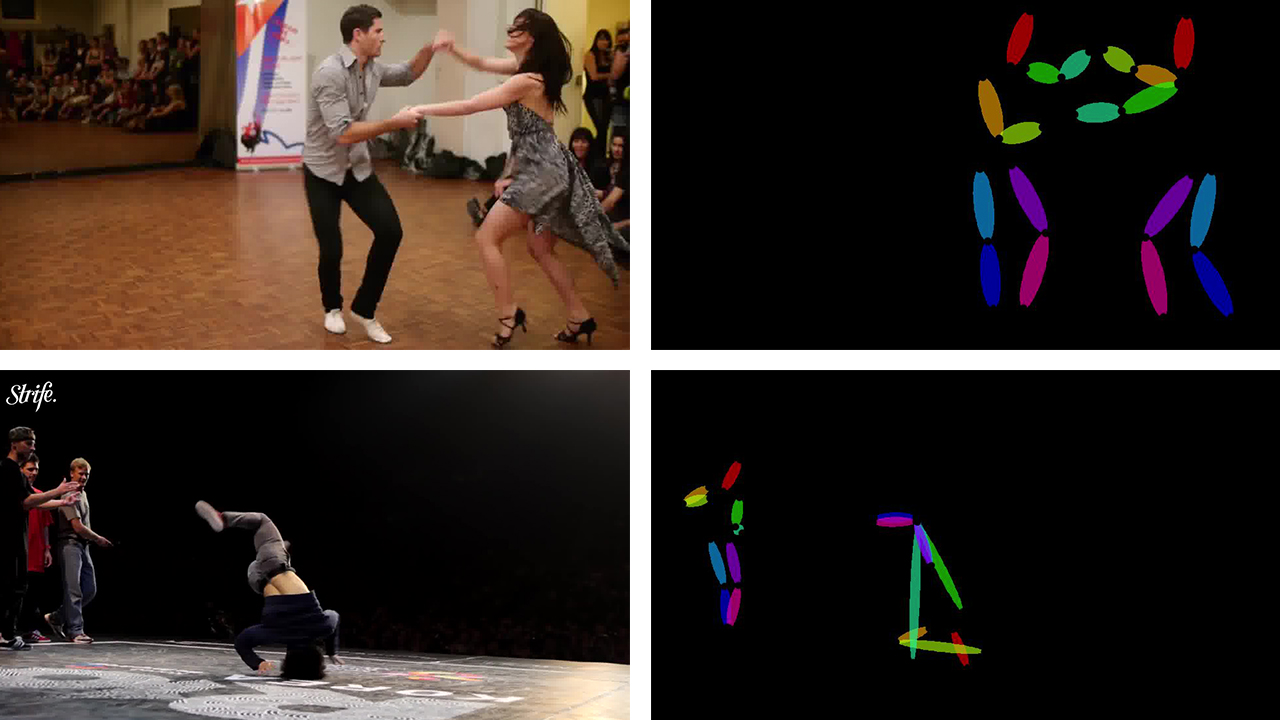}
\caption{Demonstration of outputs from our pose detection pipeline. Top: Latin dancing positively classified. Bottom: Break dancing being erroneously classified. The dancers' left leg is accurate but his remaining limbs fail.}
\label{fig:skeletons}
\end{figure}

\subsection{Temporal 3D CNN (Skeletal)} In this pipeline we compute a temporal CNN on multi-person pose information. We visualize the pipeline in Figure \ref{fig:skeletal}. This architecture demonstrates the importance of leveraging context for particular videos. Dance videos inherently benefit from this representation given that there are generally multiple people in the scene. Through the use of a visualization of pose we are able to attain comparable results to single-frame CNN approaches. It is key to note that this method does not use visual information from the original frame but solely visualized pose information as shown in Figure \ref{fig:skeletons}. Similar to our optical flow approach, it is likely that this method benefits heavily from encoding the number of people in the frame in addition to the motion over the 16 frames that are convolved in the temporal domain.

\begin{figure*}[t]
\centering
\includegraphics[width=\textwidth]{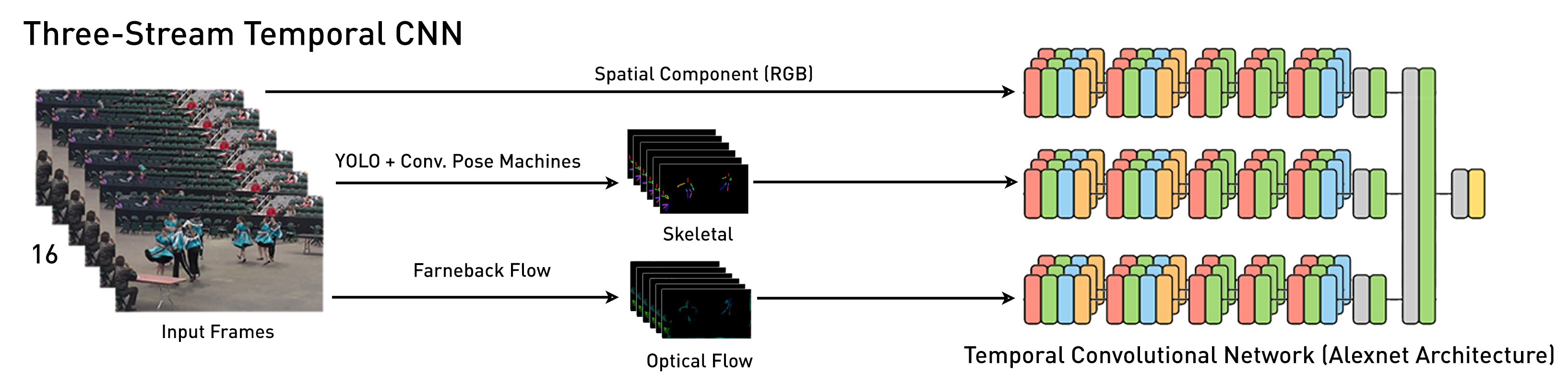}
\caption{This visualizes the workflow for our three-stream temporal CNN which uses three convolutional stacks to process the spatial and respective motion components. It aggregates the fc7 layers into one and outputs the dance classification for a 16 frame input.}
\label{fig:final}
\end{figure*}

\subsection{Three-Stream CNNs}

We tested both single-frame and temporal approaches for a three-steam convolutional network in order to directly compare the potential importance of embedding multiple frames into the learning pipeline in addition to providing multiple representations of your original input. We highlight that these temporal convolutions are computing 2D convolutions over each of the input frames. Although this increases the complexity of our model it still remains significantly more tractable than computing 3D convolutions which require approximately twice the computational power.

\subsubsection{Frame-by-Frame Architecture}

For the frame-by-frame architecture, the first stack of our network processes the spatial representation of our input which is our RGB image. Our second stack processes the optical flow representation which was computed from frames $n$ and $n-10$ in order to accentuate particular motions from a given dance. Our third stack processes our multi-person pose visualization explained in Figure \ref{fig:skeletal}. As discussed earlier, this stack is essentially encoding the number of participants detected for a given dance frame and their current pose.

\subsubsection{Temporal Architecture}

The temporal architecture utilizes the same three stacks but processes chunks of 16 frames at a time in order to incorporate a temporal component into the loss of the network. This enables us to learn motion parameters from the spatial, optical flow and multi-person pose representations. A visualization of our pipeline can be seen in Figure \ref{fig:final} whose convolutional and fully connected layers are based on the standard AlexNet architecture\cite{NIPS2012_4824}.

\section{Baseline Experiments}
We implement our proposed approaches with the goal of determining which approach is most effective at highly dynamic video classification. Implementation details for each approach are given below.

\subsection{Dataset Splits}
We extract individual frames from the Let's Dance dataset (1000 10-second videos at 30fps, resulting in 300000 frames), which we then randomly split per video into 80\% training, 10\% testing and 10\% validation (consistent across experiments). Optical flow and pose detection was split in the same manner in order to consistently test the approaches.

\subsection{Frame-by-Frame}

To perform a baseline video classification experiment, we implemented the architecture shown in Figure \ref{fig:arch1} in Tensorflow\cite{tensorflow2015-whitepaper}. The weights for the network's convolutional layers are initialized to values from a network pre-trained on the ILSVRC 2012 dataset \cite{ILSVRC15}. Final video classification results can be determined by classifying each frame in a video and voting to determine the video's overall class.

For an initial comparison, we also tested the network with optical flow imagery as the input.

Overall, we observed significant amounts of overfitting in the original training accuracy which hints at the network learning too much about the appearance of the specific videos in the training set for each class. As we hypothesized, using image frames alone results in the network learning features that do not generalize well to the dancing categories, since it has no way to observe the motion inherent in the video. Testing accuracy peaks at 56.4\% over 10,000 iterations of fine-tuning the network. We compare these results to a similar framework introduced by \cite{wang2015towards} which tested the frame-by-frame baseline on UCF-101, attaining an accuracy of 72.8\%. This directly demonstrates the possibility of solving the classification problem by carefully selecting the right frame versus understanding the underlying motion of the video.

We also ran the identical setup using optical flow estimation. Before training we pre-compute optical flow for the entire dataset. We used Farneback's method for calculating dense optical flow \cite{farneback2003two} to obtain a per-pixel estimate of the horizontal and vertical components of motion and then incorporate this into the same network architecture.

In this case we saw slightly worse performance at approximately 45\% for testing. We do note that the overfitting for optical flow images is subdued given that the per-frame images no longer contain background information. Given that a number of our dances occurred in similar or identical settings, background information was a strong confounding factor for the original images. The overall result for optical flow performs  worse than training on RGB images given that it is merely embedding the motion between two frames. We will later demonstrate that larger frame chunks provide significant improvements to this approach.

\subsection{Two-Stream Late Fusion}

We implemented the two-stream late fusion architecture shown in Figure \ref{fig:arch2} in Caffe\cite{jia2014caffe}. The two-stream approach follows intuitively from the previous subsection in which we discuss the effects of both a frame-by-frame method on images and on optical flow. Each individual stream uses the CaffeNet architecture, with weights initialized to a network pre-trained on the ILSVRC 2012 dataset \cite{ILSVRC15}. We then fine-tune the network by training only the fully-connected layers at the end of each stream, which are then concatenated and passed through a final fully-connected layer which outputs the respective classifications.

\begin{figure}[t]
\centering
\includegraphics[width=0.45\textwidth]{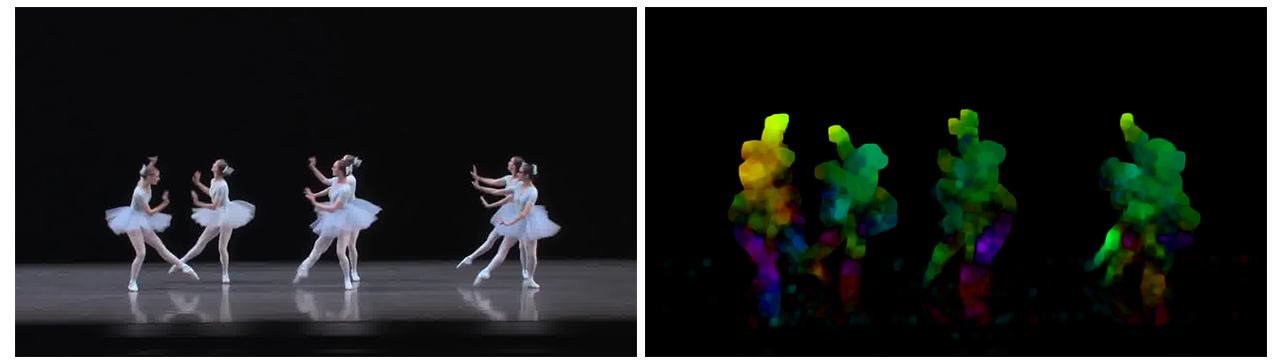}
\caption{An image of dancers performing ballet and their optical flow estimation. As we can see, optical flow does a good job of segmenting the subjects in the scene in addition to encoding their motion.}
\label{fig:optical_flow}
\end{figure}

We note that each architecture in the two-stream method is still using a single frame as input, and as such the network is trained on a frame-by-frame basis. We chose to use the CaffeNet architecture for each frame, initialized with the ILSVRC 2012 weights, to be consistent with the baseline frame-by-frame experiment described in the previous section. This allows us to perform a direct comparison between the two-stream and frame-by-frame approaches, to determine the benefit of optical flow on this dataset. As with the frame-by-frame approach, final video classifications can be determined by classifying each individual frame and optical flow image pair, followed by voting to determine an overall class. It is interesting to note that the total per video classification accuracy of this method was 68.89\% which is much higher than the single frame-by-frame accuracy of 56.40\%. Although one may be compelled to argue that single-frame motion is key to this classification, we refer back to Figure \ref{fig:skeldist}. This figure demonstrates that the frames throughout the dataset also contain a tremendously varied number of participants. As we can further see in Figure \ref{fig:optical_flow}, optical flow tends to visually separate the dancers from the background, which also explains the significant increase in the algorithm's performance. In addition to the motion in a single frame pair, the foreground's shape and representation is playing a key role in the classification of the network.

 The results demonstrate an improvement over each independent approach, with a classification accuracy of 64.69\% per-frame. This is a significant increase of 10\% above the imaging method and 20\% over the optical flow method. This increase was attained by combining the same architecture as the previous two methods, with the addition of a single concatenation node to fuse the data at the end of the network. It demonstrates that directly incorporating temporal data into a network can be immediately beneficial towards classifying video.

Leveraging the network to perform full video classification (rather than only per-frame classification), we tested the trained network on our test set of videos, taking the class with the largest number of per-frame votes as the final video label. This resulted in a per-video classification accuracy of 66.14\%. After further experimentation with the network architecture, we saw a significant improvement from computing a unique mean image to subtract from the optical flow, which increased our accuracy to a final per-video classification rate of 68.89\%.

The network performs well at classifying Ballet, Waltz, Tango, Flamenco, and Foxtrot, with poor classification accuracy on Break and Swing dancing. Of particular interest is the network's performance on Waltz, Tango and Foxtrot which occur in similar settings. As such, the network shows that it's capable of performing fine-grained classification within the Let's Dance dataset.

\begin{table}[htp!]
\begin{center}
\begin{tabular}{|c|c|c|}
\hline
\textbf{Dataset} & \textbf{Frame-by-Frame} & \textbf{Two-Stream } \\ \hline
UCF-101 \cite{soomro2012ucf101} & \textbf{72.8\%} & \textbf{88.0\%} \\ \hline
Let's Dance & 56.4\% & 68.89\% \\ \hline
\end{tabular}
\vspace{0.5em}
\caption{\label{tabucf101} Method Comparison of UCF-101 and Let's Dance. UCF Frame-by-Frame results obtained from \cite{wang2015towards}, Two-Stream results obtained from \cite{NIPS2014_5353}}
\end{center}
\end{table}

Lastly, we revisit our accuracy results with UCF-101, a well-established activity recognition dataset. Table \ref{tabucf101} illustrates high levels of accuracy on UCF-101 using the standard extensions of image classification techniques which we discuss in this section.  It is important to note that the two-stream comparison is comparing a two-stream accuracy for UCF-101 that utilizes an SVM to combine its streams whereas we concatenate the final layers of both convolutional streams into the fully connected output. As stated earlier, this illustrates the core issue we encountered in looking for a highly dynamic dataset which further validates our motivation to introduce the ``Lets Dance'' dataset to the research community.

\section{Results \& Discussion}
\label{results}

In order to assess our temporal architectures we compare with a number of state-of-the-art approaches that explicitly encode motion in order to determine their performance. Overall it has become clear to us that we need to transition from traditional per-frame CNN approaches when conducting video classification. 

It is evident from Table \ref{tab:results} that methods which embed motion significantly outperform traditional methods and that metrics to evaluate these approaches are necessary in order to better understand what each network architecture is learning.

\subsection{Temporal 3D CNN}
In order to evaluate this approach we restructured our data into 16-frame chunks that were needed as the input for the 3D convolution. The network could be trained on the 3D features from 16-frame non-overlapping chunks of the video. We fine-tuned from the network trained on UCF101 by \cite{tran2014learning}. This method yielded a per-video accuracy of 70.11\%. This result was particularly impressive because it demonstrated the inherent ability of a 3D convolution to extract motion features that are not explicitly computed. The major drawback of this approach is its complexity. A 3D convolution inherently takes significant computation for a single-stream.

We were unable to perform multi-stream approaches using 3D convolutions due to this complexity. In order to combat this we introduce more tractable approaches for state-of-the-art graphics cards (Our current systems utilizes Titan Z Pascal graphics cards) that achieve comparable performance by explicitly encoding motion into the network architecture. In addition to this we note that 3D convolutions are limited to the initial input-size which in our case was 16 frames. This makes it difficult to encode more complex motions that last more than 1/2 second without sub sampling frames which will invariably lead to a loss in detail. Most temporal methods will invariably suffer from this limitation given that variable inputs into a convolutional network has not been fully explored.

\begin{table}[t]
\begin{center}
\begin{tabular}{|c|c|}
\hline
\textbf{Approach} & \textbf{Testing Accuracy}\\ \hline
Frame-by Frame CNN & 56.4\% \\ \hline
Two-Stream CNN & 68.89\% \\ \hline
Temporal 3D CNN (RGB) & 70.11\% \\ \hline
Temporal 3D CNN (Skeletal) & 57.14\% \\ \hline
Three-Stream CNN & 69.20\% \\ \hline
Temporal Three-Stream CNN & \textbf{71.60}\% \\
\hline
\end{tabular}
\vspace{0.5em}
\caption{Comparison of numerous approaches and their testing accuracies on our dataset
\label{tab:results}}
\end{center}
\end{table}

\subsection{Skeletal Temporal 3D CNN}
In order to embed human motion data, we incorporate skeletal images into a temporal CNN. We visualize each pose into a single image which represents the pose for that particular frame. We attained an accuracy of 57.14\%. We note that this accuracy still performs marginally better than a frame-by-frame approach despite the fact that it does not utilize the spatial (RGB) representation. Due to the computational complexity of running concurrent 3D convolutional networks we propose a stacked 2D convolutional method which allows us to combine multiple streams in a single state-of-the-art graphics card.

\subsection{Frame-by-Frame Three-Stream CNN}
Our Three-Stream Frame-by-Frame architecture utilizes all three data modalities. We assess this as both a single-frame and as a stacked architecture in order to compare their benefits and drawbacks. As shown in Table \ref{tab:results}, this approach attains an accuracy of 69.20\%. This three-stream network performs comparably to the two-stream fusion approach we conducted as one of our baselines which indicates that there is not a significant amount of information added from the use of both skeletal and optical flow representations. 

\subsection{Temporal Three-Stream CNN}
Logically, we extended our frame-by-frame approach into the temporal domain by stacking the image input layers to produce a 16-frame chunk. This approach utilizes the same input as the Temporal 3D CNN we implemented at a much lower complexity for three streams. We saw this method attain the best performance out of all of the methods we evaluated, at an accuracy of 71.60\%. 

Looking at our most successful approaches, three-stream methods and 3D convolution, we note that both achieve very similar performance in per-video classification. However, the two methods are not equivalent in terms of computational resources. Beyond the increased workload and restrictions inherent in appropriately formatting the data for the temporal CNN, 3D convolution is much more computationally-intensive at both training and testing time. We observe that even though the temporal CNN was our most successful approach, it may be sub-optimal when a much simpler three-stream stacked convolutional network approach is available.

\section{Conclusion and Future Work}

In this work we sought out to understand the effect of motion on classifying videos. Recent work in the are has demonstrated the relevance of these type of videos, most recently seen in \cite{ava} and \cite{kinetics}. The work we have conducted demonstrates that traditional CNN approaches do not properly or intentionally encode motion in their methodology. This fact is frequently overlooked by testing on videos that do not inherently require motion. That was the primary motivator of this work. As we can see in Table \ref{tab:results}, 3D convolution methods outperform more traditional approaches by inherently encoding motion into their computation and prediction. Similarly, two-stream methods that incorporate optical flow can also leverage temporal features to significantly improve video classification.

This also opens up some potential for future work in incorporating optical flow and pose data. Hybrid approaches, such as a three-stream temporal CNN, have the potential to increase an algorithm's understanding of the video. We have also developed a more focused dataset that we believe the research community will benefit from by intentionally selecting highly dynamic actions in one specific class. We tested a variety of traditional and more complex methods in order to begin to understand the composition of our dataset and its baseline performance. The Let's Dance dataset will continue to help us to assess adequate motion parameterization and hopefully assist in improving how we learn from video data.

One of the biggest problems we ran into throughout this research endeavor was determining the best classes to select for our dataset. Initially we had some intuition for dancing and martial arts being adequate parent categories but we quickly saw that martial arts represented a multi-class problem. Although dancing exhibits similar overlaps the separation was much more evident when performing the data collection. We also had to alternate between different dances partly due to availability on YouTube and our own understanding of these dances.

One of the next steps we have considered in this work is modifying the input data in order to blur out regions of the video which are not in motion or considered background. This would further enforce motion parameterization and help us better understand how we can accomplish that to improve more general video classification algorithms. This could also be explored by independently classifying the pose estimation which is significantly more challenging.

We seek to extend this work by continuing to develop human-motion representations that intentionally target these highly dynamic actions. 

For more information, please visit: \url{https://www.cc.gatech.edu/cpl/projects/dance/}

{\small
\bibliographystyle{ieee}
\bibliography{egbib}
}

\end{document}